\documentclass[acmsmall]{acmart}

\usepackage{ multirow, appendix, stfloats, color, hyperref, enumitem, natbib, xcolor, array, ragged2e}

\AtBeginDocument{%
  \providecommand\BibTeX{{%
    \normalfont B\kern-0.5em{\scshape i\kern-0.25em b}\kern-0.8em\TeX}}}

\setcopyright{acmcopyright}
\copyrightyear{2022}
\acmYear{2022}
\acmDOI{10.1145/1122445.1122456}

\acmJournal{TKDD}

\begin{document}

\title{Self-Supervised Transformer for Sparse and Irregularly Sampled Multivariate Clinical Time-Series}

\author{Sindhu Tipirneni}
\email{tsaisindhura@vt.edu}
\orcid{0000-0001-5502-1616}
\affiliation{%
  \institution{Virginia Tech}
  \city{Blacksburg}
  \state{Virginia}
  \country{USA}
  \postcode{24061}
}
\author{Chandan K Reddy}
\email{reddy@cs.vt.edu}
\orcid{0000-0003-2839-3662}
\affiliation{%
  \institution{Virginia Tech}
  \streetaddress{900 N Glebe Road}
  \city{Arlington}
  \state{Virginia}
  \country{USA}
  \postcode{22203}
}


\begin{abstract}
Multivariate time-series data are frequently observed in critical care settings and are typically characterized by sparsity (missing information) and irregular time intervals. Existing approaches for learning representations in this domain handle these challenges by either aggregation or imputation of values, which in-turn suppresses the fine-grained information and adds undesirable noise/overhead into the machine learning model.
To tackle this problem, we propose a \textbf{S}elf-supervised \textbf{Tra}nsformer for \textbf{T}ime-\textbf{S}eries (STraTS) model which overcomes these pitfalls by treating time-series as a set of observation triplets instead of using the standard dense matrix representation. It employs a novel Continuous Value Embedding technique to encode continuous time and variable values without the need for discretization. It is composed of a Transformer component with multi-head attention layers which enable it to learn contextual triplet embeddings while avoiding the problems of recurrence and vanishing gradients that occur in recurrent architectures.
In addition, to tackle the problem of limited availability of labeled data (which is typically observed in many healthcare applications), STraTS utilizes self-supervision by leveraging unlabeled data to learn better representations by using time-series forecasting as an auxiliary proxy task.
Experiments on real-world multivariate clinical time-series benchmark datasets demonstrate that STraTS has better prediction performance than state-of-the-art methods for mortality prediction, especially when labeled data is limited. 
Finally, we also present an interpretable version of STraTS which can identify important measurements in the time-series data. Our data preprocessing and model implementation codes are available at \url{https://github.com/sindhura97/STraTS}.


\end{abstract}



\maketitle

\section{Introduction}
\begin{figure}
    \centering
    \includegraphics[scale=0.42, trim=0 45 0 0]{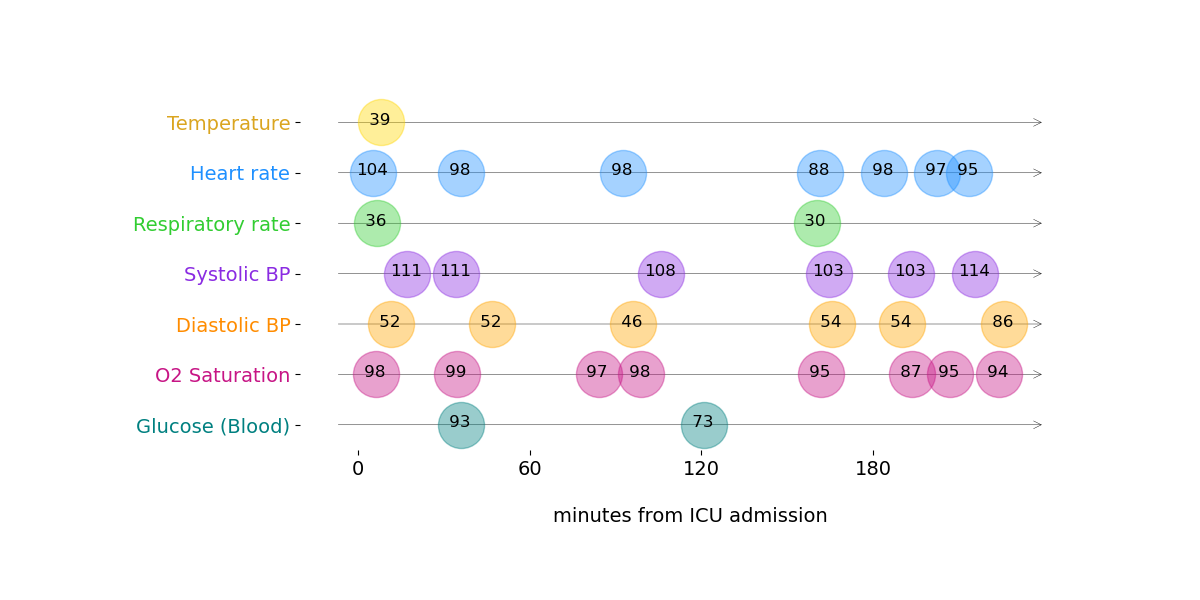}
    \caption{An illustrative example of a multivariate clinical time-series with irregular time points and missing values.}
    \label{fig:eg}
    \Description{Different variables like temperature, heart rate, etc. measured at sporadic intervals.}
\end{figure}

Time-series data is routinely collected in various healthcare settings where different measurements are recorded for patients throughout their course of stay (See Figure \ref{fig:eg} for an illustrative example). Predicting clinical outcomes like mortality, decompensation, length of stay, and disease risk from such complex multivariate time-series data can facilitate both effective management of critical care units and automatic personalized treatment recommendation for patients. The success of deep learning in image and text domains realized by convolutional and recurrent networks \citep{sutskever2014sequence, chung2014empirical}, and Transformer models \citep{vaswani2017attention} have inspired the application of these architectures to develop better prediction models for time-series data as well. However, time-series in the clinical domain portray a unique set of challenges that are described below. 
\begin{itemize}
    \item \textbf{Missingness and Sparsity}: A patient's condition may demand observing only a subset of variables of interest. Thus, not all the variables are observed for every patient. Also, the observed time-series matrices are very sparse as some variables may be measured more frequently than others for a given patient.
    \item \textbf{Irregular time intervals  and Sporadicity}: Not all clinical variables are measured at regular time intervals. Thus, the measurements may occur sporadically in time depending on the underlying condition of the patient.
    \item  \textbf{Limited labeled data}: Patient-level clinical data is often expensive to obtain and labeled data subsets pertaining to a specific prediction task may be even more limited (for e.g., building a severity classifier for Covid-19 patients.)
\end{itemize}

\label{sec:intro}

A straight-forward approach to deal with irregular time intervals and missingness is to aggregate measurements into discrete time intervals and add missingness indicators, respectively. However, this suppresses important fine-grained information because the granularity of observed time-series may differ from patient to patient based on the underlying medical condition.
Existing sequence models for clinical time-series \citep{che2018recurrent} and other interpolation-based models \citep{shukla2019interpolation} address this issue by including a learnable imputation or interpolation component.
Such techniques add undesirable noise and extra overhead to the model which usually worsens as the time-series become increasingly sparse. These models rely on an effective imputation/interpolation scheme in order to achieve strong performance on the target task. But it is unreasonable to impute clinical variables without careful consideration of the domain knowledge about each variable which might be non-trivial to obtain.

Considering these shortcomings, we design a framework that does not need to perform any such operations and directly builds a model \textit{based only on the observations that are available in the data}.
Thus, unlike conventional approaches which view each time-series as a matrix of certain dimensions (\#features $\times$ \#time-steps), our model regards each time-series as a set of observation triplets (a triple containing time, variable, and value) without the necessity for aggregation or imputation. The proposed STraTS (acronym for \textbf{S}elf-supervised \textbf{Tra}nsformer for \textbf{T}ime-\textbf{S}eries) model embeds these triplets by using a novel Continuous Value Embedding (CVE) scheme to avoid the need for binning continuous values before embedding them. The use of CVE for representing the time dimension preserves the fine grained information which is lost when the time-axis is discretized. STraTS encodes contextual information of observation triplets using a Transformer-based architecture with multi-head attention. We choose this over recurrent neural network (RNN) architectures because the sequential nature of RNN models hinders parallel processing while the Transformer bypasses this by using self-attention to attend from every token to every other token in a single step.

To build robust representations using limited labeled data, we employ self-supervision and develop a time-series forecasting task to pretrain STraTS. This enables learning generalized representations in the presence of limited labeled data and alleviates sensitivity to noise. Furthermore, interpretable models are usually preferred in healthcare but existing deep models for clinical time-series lack this component. Thus, we also propose an interpretable version of our model (I-STraTS) which slightly compromises on performance metrics but can identify important measurements in the input. Though we evaluate the proposed model only on binary classification tasks, our framework can also be utilized in other supervised and unsupervised settings, where learning robust and generalized representations of sparse and sporadic time-series is desired. The main contributions of our work can be summarized as follows.

\begin{itemize}
    \item Propose a Transformer-based architecture called STraTS for clinical time-series which addresses the unique challenges of missingness and sporadicity of such data by avoiding aggregation and imputation.
    \item We Develop a novel Continuous Value Embedding (CVE) mechanism using a one-to-many feed-forward network to embed continuous times and measured values in order to preserve fine grained information.
    \item Utilize forecasting as a self-supervision (proxy) task to leverage unlabeled data to learn more generalized and robust representations.
    \item Propose an interpretable version of STraTS that can be used when interpretability is more desired compared to quantitative performance gains. 
    \item Demonstrate through an extensive set of experiments that the design choices of STraTS lead to a better performance compared to competitive baseline models for mortality prediction on two real-world clinical datasets.
\end{itemize}

The rest of this paper is organized as follows. In Section \ref{sec:related}, we review relevant literature about tackling sparse and sporadic time-series data, and self-supervised learning. Section \ref{sec:proposed} formally defines the prediction problem and gives a detailed description of the architecture of STraTS along with the self-supervision approach. Section \ref{sec:exp} presents experimental results comparing STraTS with various baselines and demonstrates the interpretability of I-STraTS with a case study. Finally, Section 5 concludes the paper and provides future directions.

\section{Related Work}
\label{sec:related}
\subsection{Clinical Time-Series}

A straightforward approach to address missing values and irregular time intervals is to impute and aggregate the time-series, respectively, before feeding them to a classifier \cite{lipton2015phenotyping, chen2018dynamic}. However, such classifiers ignore the missingness in the data which can be quite informative.
\citeauthor{lipton2016directly} \cite{lipton2016directly} show that phenotyping performance can be improved by passing missingness indicators as additional features to an RNN classifier. But they still lose fine-grained information by aggregating each time-series into hourly intervals.

Several early works rely on Gaussian Processes (GP) \citep{rasmussen2003gaussian} to model irregular time-series. For example, \citeauthor{lu2008reproducing} \cite{lu2008reproducing} represent each time-series as a smooth curve in a reproducing kernel Hilbert space (RKHS) using GP by optimizing GP parameters using Expectation Maximization (EM), and then derive a distance measure on the RKHS which is used to define the SVM classifier's kernel. To account for uncertainty in GP, \citeauthor{li2015classification} \cite{li2015classification} formulate the kernel by applying an uncertainty-aware base kernel (called the expected Gaussian kernel) to a series of sliding windows. These works take a two-step approach by first optimizing GP parameters and then training the classification model.
To enable end-to-end training, \citeauthor{li2016scalable} \cite{li2016scalable} again represent time-series using GP posterior at predefined time points but
use the reparametrization trick to back-propagate the gradients through a black-box classifier (learnable by gradient-descent) into the GP model. The end-to-end model is uncertainty-aware as the output is formulated as a random variable. \citeauthor{futoma2017learning} \cite{futoma2017learning} extend this idea to multivariate time-series with the help of multitask GP \citep{bonilla2008multi} to consider inter-variable similarities. Though Gaussian Processes provide a systematic way to deal with uncertainty, they are expensive to learn and their flexibility is limited by the choice of covariance and mean functions.

\citeauthor{shukla2019interpolation} \cite{shukla2019interpolation} also propose an end-to-end method that constitutes interpolation and classification networks stacked in a sequence. They develop learnable interpolation layers to approximate the time-series at regular predefined time points in a deterministic fashion (unlike GP-based methods) and allow information sharing across both time and variable dimensions. However, the input to the classifier is a densely interpolated multivariate time-series which causes loss of information if the number of interpolation points is small and slows down computations while adding noise otherwise.

Instead of using a separate interpolation module followed by a traditional classifier, other approaches modify traditional recurrent architectures for clinical time-series to deal with missing values and/or irregular time intervals. For example, \citeauthor{baytas2017patient} \cite{baytas2017patient} developed a time-aware long-short term memory (T-LSTM) which is a modification of the LSTM cell to adjust the hidden state according to the irregular time gaps. ODE-RNN \citep{rubanova2019latent} uses ODEs to model the continuous-time dynamics of the hidden state while also updating the hidden state at each observed time point using a standard GRU cell. 
The GRU-D model \citep{che2018recurrent} is a modification of the GRU cell which decays inputs (to global means) and hidden states through unobserved time intervals. 
DATA-GRU \citep{tan2020data}, in addition to decaying the GRU hidden state according to elapsed time, also employs a dual attention mechanism based on missingness and imputation reliability to process inputs before feeding them to a GRU cell. All these methods use an RNN with sequence length being the number of unique timestamps in the input, which can be quite large for irregular time-series, and as a result, can slow down computations.

The imputation/interpolation schemes in the models discussed above can lead to excessive computations and unnecessary noise particularly when missing rates are quite high. Our model is designed to circumvent this issue by representing sparse and irregular time-series as a set of observations. \citeauthor{horn2020set} \cite{horn2020set} develop SeFT with a similar idea and use a parametrized set function for classification. The attention-based aggregation used in SeFT contains the same queries for all observations to facilitate low memory and time complexity while compromising on accuracy. The initial embedding in SeFT contains fixed time encodings while our approach uses learnable embeddings for all the three components (time, variable, value) of the observation triplet.

The challenge of training in scenarios with limited labeled data still remains.
In order to address this issue, we turn towards self-supervision for a better utilization of the available data to learn effective representations.


\subsection{Self-supervised Learning}
Supervised deep learning models often rely on large amounts of labeled data to learn generalized and robust representations. Limited labeled data can make the model easily overfit to training data and make the model more sensitive to noise. Since labeled data is expensive to obtain, self-supervised learning was introduced as a technique to solve this challenge. This technique trains the model on carefully constructed proxy tasks that improve the model's performance on target prediction tasks. The labeled datasets for proxy tasks are obtained from the unlabeled data in an inexpensive semi-automatic process. 
Yann Le Cunn\footnote{\url{https://drive.google.com/file/d/1r-mDL4IX_hzZLDBKp8_e8VZqD7fOzBkF/view}} describes self-supervised learning as to "predict any part of the input from any
other part". Self-supervised learning enables the model to learn correlations in input data which enhance the model's learning of supervised target prediction tasks.
\citeauthor{liu2021self} \cite{liu2021self} review the state-of-the-art self-supervised learning methods in computer vision, natural language processing, and graph representation learning. 
Though this technique has shown great performance boosts with image \citep{jing2020self} and text \citep{devlin2018bert, yang2019xlnet} data, its application to time-series data has been limited. One such effort is made by \citeauthor{jawed2020self} \cite{jawed2020self} which uses a 1D CNN for dense univariate time-series classification and shows increased accuracy by using forecasting as an additional task in a muti-task learning framework. 
\citeauthor{zerveas2021transformer} \cite{zerveas2021transformer} pretrained a Transformer model using a denoisining objective and showed improved performance on regression and classification tasks with dense multivariate time-series.
In our work, we demonstrate time-series forecasting as a viable and effective self-supervision task for a Transformer model. Our work is the first to explore self-supervised learning in the context of sparse and irregular multivariate time-series.



\section{Proposed Approach}
\label{sec:proposed}
In this section, we describe our STraTS model by first introducing the problem with relevant notation and definitions and then explaining the different components of the model which are illustrated in Figure \ref{fig:arc}.

\begin{figure}[ht]
    \centering
    \includegraphics[scale=0.7, trim=100 0 100 0]{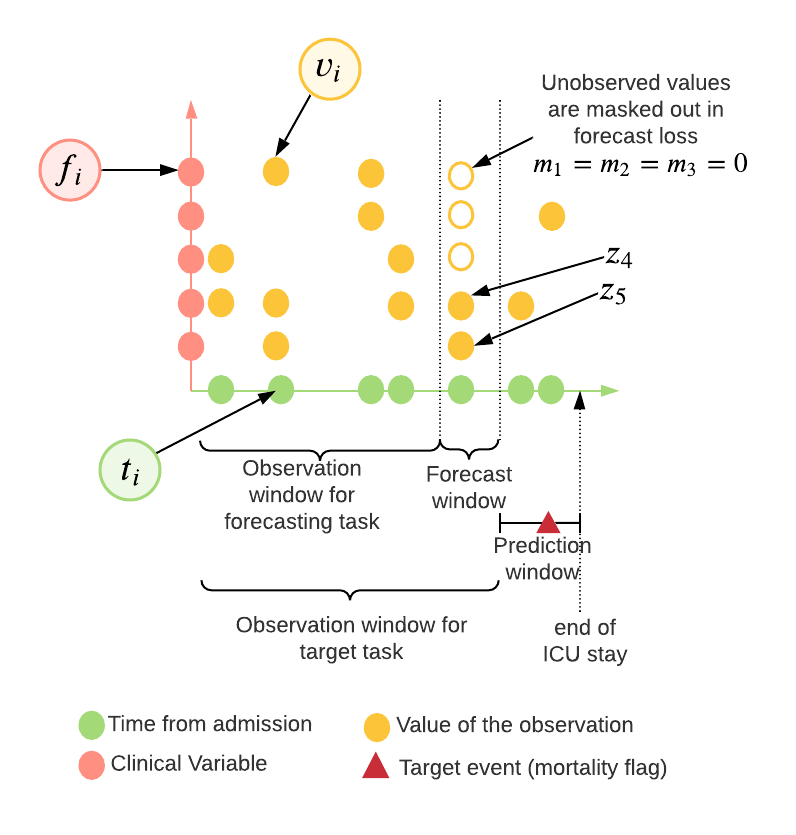}
    \caption{An illustration of input and output construction for target and self-supervision (forecasting) tasks. The target task uses a fixed length observation window to predict in-hospital mortality. The forecasting task has an observation window that is followed by a fixed length prediction window in which only a subset of variables may be observed.
    Note that several observation windows are considered for each time-series for the forecasting task.}
    \label{fig:io}
    \Description{A sparse and sporadic multivariate time series marked with an observation window followed by forecast/prediction window. There may be unobserved variables in the forecast window which are masked out in forecast loss.}
\end{figure}

\subsection{Problem Definition}
As stated in the previous sections, STraTS represents each time-series as a set of observation triplets. Formally, an \textit{observation triplet} is defined as a triple $(t,f,v)$ where $t\in \mathbb{R}_{\geq 0}$ is the time, $f\in\mathcal{F}$ is the feature/variable, and $v\in \mathbb{R}$ is the value of the observation. A \textit{multivariate time-series} $\mathbf{T}$ of length $n$ is a defined as a set of $n$ observation triplets i.e., $\mathbf{T}=\{(t_i,f_i,v_i)\}_{i=1}^n$.

Consider a dataset $\mathcal{D}=\{(\mathbf{d}^k, \mathbf{T}^k, y^k)\}_{k=1}^N$ with $N$ labeled samples, where the $k^{th}$ sample contains a demographic vector $\mathbf{d}^k\in\mathbb{R}^{D}$, a multivariate time-series $\mathbf{T}^k$, and a corresponding binary label $y^k\in\{0,1\}$.
In this work, each sample corresponds to a single ICU stay where several clinical variables of the patient are measured at irregular time intervals and the binary label indicates in-hospital mortality. 
The underlying set of time-series variables denoted by $\mathcal{F}$ may include vitals (such as temperature), lab measurements (such as hemoglobin), and input/output events (such as fluid intake and urine output). Thus, the \textit{target task} aims to predict $y^k$ given $(\mathbf{d}^k, \mathbf{T}^k)$.

Our model also incorporates forecasting as a self-supervision task. For this task, we consider a bigger dataset with $N'\geq N$ samples given by $\mathcal{D'}=\{(\mathbf{d}^k, \mathbf{T}^k, \mathbf{m}^k, \mathbf{z}^k)\}_{k=1}^{N'}$. Here, $\mathbf{m}^k\in \{0,1\}^{|\mathcal{F}|}$ is the forecast mask which indicates whether each variable was observed in the forecast window and  $\mathbf{z}^k\in\mathbb{R}^{|\mathcal{F}|}$ contains the corresponding variable values when observed. The forecast mask is necessary because the unobserved forecasts cannot be used in training and are hence masked out in the loss function.
The time-series in this dataset are obtained from both the labeled and unlabeled time-series by considering different observation windows. Figure \ref{fig:io} illustrates the construction of inputs and outputs for the target task and forecasting task.


\begin{table}[]
\centering
    \caption{Notations used in this paper.}
    \begin{tabular}{cl}
    \toprule
    Notation & Definition \\
    \midrule
    $N$ & \# Time-series for target task \\
    $N'$ & \# Time-series for forecasting task \\
    $\mathbf{d}\in\mathbb{R}^{D}$ & Demographics vector \\
       $\mathcal{F}$  &Set of clinical variables  \\
       $t_i\in \mathbb{R}_{\geq 0}$ &Time of $i^{th}$ observation \\
       $f_i\in\mathcal{F}$ &Variable of $i^{th}$ observation \\
       $v_i\in \mathbb{R}$ &Value of $i^{th}$ observation \\
       $(t,f,v)$ &Observation triplet \\ 
        $\mathbf{T}=\{(t_i,f_i,v_i)\}_{i=1}^n$ & Multivariate time-series \\
        $y,\tilde{y}\in \{0,1\}$ &True and predicted outputs for target task \\
        $\mathbf{z}, \tilde{\mathbf{z}}\in\mathbb{R}^{|\mathcal{F}|}$ & True and predicted outputs for forecasting task \\
        $\mathbf{m}\in\{0,1\}^{|\mathcal{F}|}$ & Forecast mask\\
        $\mathbf{e}^t_i,\,\mathbf{e}^v_i\in \mathbb{R}^d$ &CVE for time and value \\
        $\mathbf{e}^f_i\in \mathbb{R}^d$ &Variable embedding \\
        $\mathbf{e}_i\in \mathbb{R}^d$ &Initial triplet embedding \\
        $\mathbf{e}^T\in \mathbb{R}^d$ &Time-series embedding \\
        $\mathbf{e}^d\in \mathbb{R}^d$ &Demographics embedding \\
    \bottomrule
    \end{tabular}
\label{tab:not}
\end{table}

\subsection{Architecture of STraTS}
\begin{figure*}[ht]
    \centering
    \includegraphics[scale=0.545, trim=14 0 0 0]{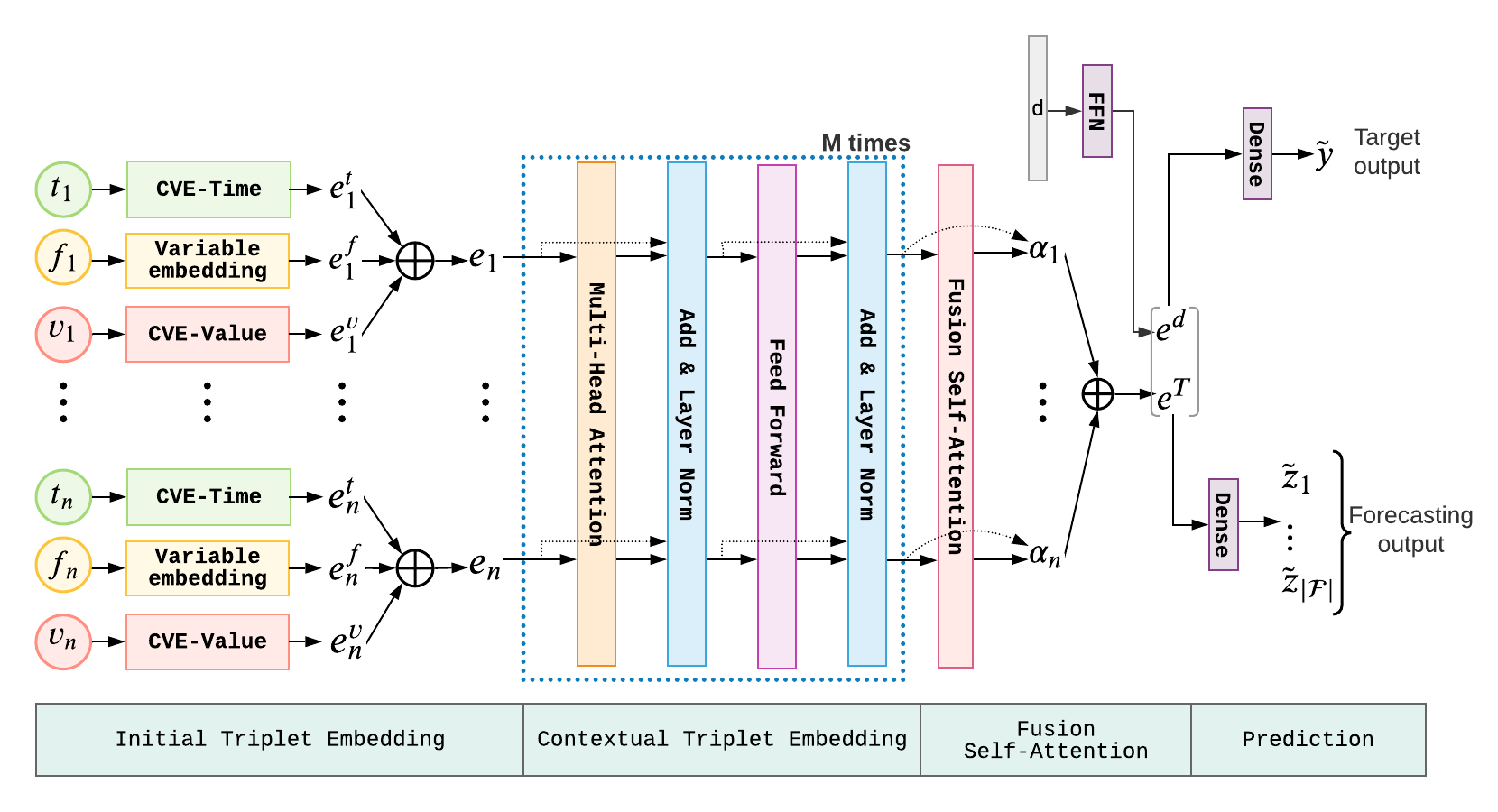}
    \caption{The overall architecture of the proposed STraTS model. The Input Triplet Embedding module embeds each observation triplet, the Contextual Triplet Embedding module encodes contextual information for the triplets, the Fusion Self-Attention module computes time-series embedding which is concatenated with demographics embedding and passed through a dense layer to generate predictions for target and self-supervision (forecasting) tasks. }
    \label{fig:arc}
    \Description{Fully described in text.}
\end{figure*}

The architecture of STraTS is illustrated in Figure \ref{fig:arc}.
Unlike most of the existing approaches which take a time-series matrix as input, STraTS \textit{defines its input as a set of observation triplets}. 
Each observation triplet in the input is embedded using the Initial Triplet Embedding module. The initial triplet embeddings are then passed through a Contextual Triplet Embedding module which utilizes the Transfomer architecture to encode the context for each triplet. The Fusion Self-attention module then combines these contextual embeddings via self-attention mechanism to generate an embedding for the input time-series which is concatenated with demographics embedding and passed through a feed-forward network to make the final prediction. The notations used in the paper are summarized in Table \ref{tab:not}.

\subsubsection{Initial Triplet Embedding}Given an input time-series $\mathbf{T}=\{(t_i,f_i,v_i)\}_{i=1}^n$, the initial embedding for the $i^{th}$ triplet $\mathbf{e_i}\in\mathbb{R}^d$ is computed by summing the following component embeddings: (i)
\textit{Feature embedding }$\mathbf{e^f_i}\in\mathbb{R}^d$,
 (ii) \textit{Value embedding} $\mathbf{e^v_i}\in\mathbb{R}^d$, and (iii)
\textit{ Time embedding }$\mathbf{e^t_i}\in\mathbb{R}^d$. In other words, $\mathbf{e_i} = \mathbf{e^f_i} + \mathbf{e^v_i} + \mathbf{e^t_i}\;\in \mathbb{R}^d$.
Feature embeddings $\mathbf{e^f(\cdot)}$ are obtained from a simple lookup table similar to word embeddings.
Since feature values and times are continuous unlike feature names which are categorical objects, we cannot use a lookup table to embed these continuous values unless they are categorized. Some researchers \citep{vaswani2017attention, yin2020identifying} have used sinusoidal encodings to embed continuous values. We propose a novel continuous value embedding (CVE) technique using a one-to-many Feed-Forward Network (FFN) with learnable parameters i.e.,
$\mathbf{e^v_i}=FFN^v(v_i)$, and $\mathbf{e^t_i}=FFN^t(t_i)$.

Both FFNs have one input neuron and $d$ output neurons and a single hidden layer with $\lfloor\sqrt{d}\rfloor$ neurons and $tanh(\cdot)$ activation. They are of the form $FFN(x)=U\,tanh(Wx+b)$ where the dimensions of weights $\{W,b,U\}$ can be inferred from the size of hidden and output layers of the FFN.
Unlike sinusoidal encodings with fixed frequencies, this technique offers more flexibility by allowing end-to-end learning of continuous value and time embeddings without the need to categorize them.

\subsubsection{Contextual Triplet Embedding}
The initial triplet embeddings $\{\mathbf{e}_1,...,\mathbf{e}_n\}$ are then passed through a Transformer architecture \citep{vaswani2017attention} with $M$ blocks, each containing a Multi-Head Attention (MHA) layer with $h$ attention heads and a FFN with one hidden layer. Each block takes $n$ input embeddings $\mathbf{E} \in \mathbb{R}^{n\times d}$ and outputs the corresponding $n$ output embeddings $\mathbf{C}\in \mathbb{R}^{n\times d}$ that capture the contextual information. MHA layers use multiple attention heads to attend to information contained in different embedding projections in parallel. The computations of the MHA layer are given by
\begin{align}
        &\mathbf{H}_j = softmax\bigg(\frac{(\mathbf{EW}^q_j)(\mathbf{EW}^k_j)^T}{\sqrt{d/h}}\bigg)(\mathbf{EW}^v_j)\quad j=1,...,h\\
    &MHA(\mathbf{E}) = (\mathbf{H}_1 \circ ... \circ \mathbf{H}_h)\,\mathbf{W}_c 
\end{align}
Each head projects the input embeddings into query, key, and value subspaces using matrices $\{\mathbf{W}^q_j, \mathbf{W}^k_j, \mathbf{W}^v_j\} \subset \mathbb{R}^{d\times d_h}$. The queries and keys are then used to compute the attention weights which are used to compute weighted averages of value (different from value in observation triplet) vectors. Finally, the outputs of all heads are concatenated and projected to original dimension with $\mathbf{W}_c\in \mathbb{R}^{hd_h \times d}$.
The FFN layer takes the form
\begin{align}
    \mathbf{F}(\mathbf{X}) &= ReLU(\mathbf{X}\mathbf{W}^f_1 + \mathbf{b}^f_1)\,\mathbf{W}^f_2 + \mathbf{b}^f_2
\end{align}
with weights $\mathbf{W^f_1} \in \mathbb{R}^{d\times 2d},\, \mathbf{b_1^f}\in \mathbb{R}^{2d},\, \mathbf{W^f_2}\in \mathbb{R}^{2d\times d},\, \mathbf{b_2^f}\in \mathbb{R}^{d}$. 
Dropout, residual connections, and layer normalization are added for every MHA and FFN layer. Also, attention dropout randomly masks out some positions in the attention matrix before the softmax computation during training.
The output of each block is fed as input to the succeeding one, and the output of the last block gives the contextual triplet embeddings $\{\mathbf{c}_1,...,\mathbf{c}_n\}$.

\subsubsection{Fusion Self-attention}
After computing contextual embeddings using a Transformer, we fuse them using a self-attention layer to compute time-series embedding $\mathbf{e}^T\in \mathbb{R}^d$. This layer first computes attention weights $\{\alpha_1,...,\alpha_n\}$ by passing each contextual embedding through a FFN and computing a softmax over all the FFN outputs.
\begin{align}
    a_i &= \mathbf{u}_a^T\,tanh(\mathbf{W}_a\mathbf{c}_i+\mathbf{b}_a)\\
    \alpha_i &= \frac{exp(a_i)}{\sum_{j=1}^nexp(a_j)}  \quad \forall i=1,...,n
\end{align}
$\mathbf{W}_a\in \mathbb{R}^{d_a \times d}, \mathbf{b}_a\in\mathbb{R}^{d_a}, \mathbf{u_a}\in\mathbb{R}^{d_a}$ are the weights of this attention network which has $d_a$ neurons in the hidden layer. The time-series embedding is then computed as 
\begin{align}
    \mathbf{e}^T = \sum_{i=1}^n\alpha_i\mathbf{c_i}
\end{align}

\subsubsection{Demographics Embedding}
We realize that demographics can be encoded as triplets with a default value for time. However, we found that the prediction models performed better in our experiments when demographics are processed separately by passing $\mathbf{d}$ through a FFN as shown below. The demographics embedding is thus obtained as 
\begin{align}
    \mathbf{e}^d = tanh(\mathbf{W}^d_2 \,tanh(\mathbf{W}^d_1 \mathbf{d} + \mathbf{b}^d_1) + \mathbf{b}^d_2) \in \mathbb{R}^d
\end{align}
where the hidden layer has a dimension of $2d$.

\subsubsection{Prediction Head}
The final prediction for target task is obtained by passing the concatenation of demographics and time-series embeddings through a dense layer with weights $\mathbf{w}_o^T\in \mathbb{R}^d$, $b_o\in \mathbb{R}$ and  sigmoid activation.
\begin{align}
    \tilde{y} = sigmoid(\mathbf{w}_o^T[\mathbf{e}^d \circ \mathbf{e}^T]+b_o)
\end{align}
The model is trained on the target task using cross-entropy loss.

\subsubsection{Self-supervision}
We experimented with both masking and forecasting as pretext tasks for providing self-supervision and found that forecasting improved the results on target tasks. 
The forecasting task uses the same architecture as the target task except for the prediction layer i.e.,
\begin{align}
    \tilde{\mathbf{z}} = \mathbf{W}_s[\mathbf{e}^d \circ \mathbf{e}^T]+\mathbf{b}_s \in \mathbb{R}^{|\mathcal{F}|}
\end{align}
A masked MSE loss is used for training on the forecasting task to account for missing values in the forecast outputs. 
Thus, the loss for self-supervision is given by 
\begin{align}
    \mathcal{L}_{ss} = \frac{1}{|N'|} \sum_{k=1}^{N'} \sum_{j=1}^{|\mathcal{F}|} m^k_j(\tilde{\mathbf{z}}^k_j - \mathbf{z}^k_j)^2
\end{align}
where $\mathbf{m}^k_j=1$ (or $\mathbf{m}^k_j=0$) if the ground truth forecast $\mathbf{z}^k_j$ is available (or unavailable) for $j^{th}$ variable in $k^{th}$ sample. The model is first pretrained on the self-supervision task and is then fine-tuned on the target task.

\subsection{Interpretability}
We also propose an interpretable version of our model which we refer to as I-STraTS. Inspired by \citeauthor{choi2016retain} \cite{choi2016retain} and \citeauthor{zhang2020inprem} \cite{zhang2020inprem}, we alter the architecture of STraTS in such a way that the output can be expressed using a linear combination of components that are derived from individual features. Specifically, the output of I-STraTS is formulated as
\begin{align}
    \tilde{y} = sigmoid\bigg(\mathbf{w}_o^T\Big[\mathbf{d} \circ \sum_{i=1}^n \alpha_i\mathbf{e}_i \Big] + b_o\bigg)
\end{align}
Contrary to STraTS, (i) we combine the initial triplet embeddings using the attention weights in Fusion Self-attention module, and
(ii) directly use the raw demographics vector as the demographics embedding.
The above equation can also be written as 
\begin{align}
    \tilde{y} = sigmoid\bigg(\sum_{j=1}^D \mathbf{w}_o[j]\,\mathbf{d}[j] + \sum_{i=1}^n \sum_{j=1}^d \alpha_i\, \mathbf{w}_o[j+D]\,\mathbf{e}_i[j]  + b_o\bigg)
\end{align}
Thus, we assign a `contribution score' to the $j^{th}$ demographic feature as $\mathbf{w}_o[j]\mathbf{d}[j]$ and to the $i^{th}$ time-series observation as $ \sum_{j=1}^d \alpha_i\, \mathbf{w}_o[j+D]\,\mathbf{e}_i[j]$.

\section{Experiments}
We evaluated our proposed STraTS model against state-of-the-art baselines on two real-world EHR databases for the mortality prediction task. This section starts with a description of the datasets and baselines, followed by a discussion of results focusing on generalization and interpretability.
\begin{table}
\centering
    \caption{Basic statistics of the two datasets used in our experiemnts. Note that the Avg. variable missing rate and Avg. \# observations/stay are calculated using only the supervised samples.}
    \label{tab:dataset_stats}
    \begin{tabular}{lcc}
    \toprule
    &MIMIC-III &PhysioNet-2012 \\
    \midrule
     \# ICU stays &52,871 &11,988 \\
     \# ICU stays (supervised) &44,812 &11,988 \\
     \# Avg. span of time-series &101.9h &47.3h \\
     \# Avg. span of time-sries (supervised) &23.5h &47.3h\\
      \# Variables &129 &37\\
      Avg. variable missing rate &89.7\% &79.7\%  \\
      Avg. \# observations/stay &401 &436\\
      Demographics &Age, Gender &Age, Gender, Height, ICU Type \\
      Task &24-hour mortality &48-hour mortality \\
      \% positive class &9.7\% &14.2\% \\
      \bottomrule
    \end{tabular}
\end{table}
\label{sec:exp}
\subsection{Datasets}
We experiment with time-series extracted from two real-world EHR datasets which are described below. 
The dataset statistics are summarized in Table \ref{tab:dataset_stats}.  \\ \\
\noindent
\textbf{MIMIC-III} \citep{mimiciii}: This is a publicly available database containing medical records of about $46k$ critical care patients in Beth Israel Deaconess Medical Center between 2001 and 2012. We filtered ICU stays to include only adult patients and extracted  $129$ features from the following tables: input events, output events, lab events, chart events, and prescriptions for each ICU stay. For mortality prediction task, we only include ICU stays that lasted for atleast one day with the patient alive at the end of first day, and predict in-hospital mortality using the first $24$ hours of data. For forecasting, the set of observation windows is defined (in hours) as $\{ [min(0, x-24), x) \;|\; 20\leq x\leq 124,\; x\%4=0\}$ and the prediction window is the $2$-hour period following the observation window. Note that we only consider those samples which have atleast one time-series measurement in both observation and prediction windows. The data is split at patient level into training, validation, and test sets in the ratio $64:16:20$.
\\ 

\noindent
\textbf{PhysioNet Challenge 2012} \citep{goldberger2000physiobank}: This processed dataset from Physionet Challenge 2012 \footnote{\url{https://physionet.org/content/challenge-2012/1.0.0/}} contains records of $11,988$  ICU stays of adult patients. 
The target task aims to predict in-hospital mortality given the first $48$ hours of data for each ICU stay. Since demographic variables `gender' and `height' are not available for all ICU stays, we perform mean imputation and add missingness indicators for them as additional demographic variables.
To generate inputs and outputs for forecasting, the set of observation windows is defined (in hours) as $\{[0, x)\,|\;12\leq x\leq 44,\,x\%4=0\}$ and the prediction window is the $2$-hour period following the observation window. The data from set-b and set-c together is split into training and validation (80:20) while set-a is used for testing.

\subsection{Baseline Methods} 
To demonstrate the effectiveness of STraTS over the state-of-the-art methods, 
we compare it with the following baseline models.
\begin{itemize}
    \item \textbf{Gated Recurrent Unit (GRU)} \citep{chung2014empirical}: The input is a time-series matrix with hourly aggregation where missing variables are mean-imputed. Binary missingness indicators and time 
    since the last observation of each variable are also included as additional features at each time step. The final hidden state is transformed by a dense layer to generate output.
    \item \textbf{Temporal Convolutional Network (TCN)} \citep{bai2018empirical}: This model takes the same input as GRU which is passed through a stack of temporal convolution layers with residual connections. The representation from the last time step of the last layer is transformed by a dense layer to generate output.
    \item \textbf{Simply Attend and Diagnose (SaND)} \citep{song2018attend}: This model also has the same input representation as GRU and the input is passed through a Transformer with causal attention and a dense interpolation layer.
    \item \textbf{GRU with trainable Decays (GRU-D)} \citep{che2018recurrent}: The GRU-D cell takes a vector of variable values at each time one or more measurements are seen. The GRU-D cell, which is a modification to the GRU cell, decays unobserved values in this vector to global mean values and also adjusts the hidden state according to elapsed times since the last observation of each variable.
    \item \textbf{Interpolation-prediction Network (InterpNet)} \citep{shukla2019interpolation}: This model consists of a semi-parametric interpolation network that interpolates all variables at regular predefined time points,
    followed by a prediction network which is a GRU. It also uses a reconstruction loss to enhance the interpolation network. The input representation is similar to that of GRU-D and therefore, no aggregation is performed.
    \item \textbf{Set Functions for Time Series (SeFT)} \citep{horn2020set}: This model also inputs a set of observation triplets, similar to STraTS. It uses sinusoidal encodings to embed times and the deep network used to combine the observation embeddings is formulated as a set function using a simpler but faster variation of multi-head attention.
\end{itemize}
For all the baselines, we use two dense layers to get the demographics encoding and concatenate it to the time-series representation before the last dense layer. All the baselines use sigmoid activation at the last dense layer for mortality prediction. The time-series measurements (by variable) and demographics vectors are normalized to have zero mean and unit variance. All models are trained using the Adam optimizer \citep{kingma2014adam}.

\subsection{Evaluation Metrics}
The following metrics are used to quantitatively compare the baselines and proposed models for the binary classification task of mortality prediction. (i) ROC-AUC: Area under ROC curve. (ii) PR-AUC: Area under precision-recall curve. (iii) min(Re, Pr): This metric is computed as the maximum of ‘minimum of recall and precision’ across all thresholds.

\begin{table}[h]
\small
    \caption{Hyperparameters used for our experiments in this paper.}
    \begin{tabular}{p{2cm}@{\hskip 1cm}p{4.5cm}@{\hskip 1cm}p{4.5cm}}
    \toprule
    Model &MIMIC-III &PhysioNet-2012  \\
    \midrule
    GRU  &units=50, rec d/o=0.2, output d/o=0.2, lr=0.0001 &units=43, rec d/o=0.2, output d/o=0.2, lr=0.0001\\ 
    \hline
    TCN &layers=4, filters=128, kernel size=4, d/o=0.1, lr=0.0001 &layers=6, filters=64, kernel size=4, d/o=0.1, lr=0.0005\\
    \hline
    SAnD &N=4, r=24, M=12, d/o=0.3, d=64, h=2, he=8, lr=0.0005 &N=4, r=24, M=12, d/o=0.3, d=64, h=2, he=8, lr=0.0005\\
    \hline
    GRU-D  &units=60, rec d/o=0.2, output d/o=0.2, lr=0.0001 &units=49  rec d/o=0.2, output d/o=0.2, lr=0.0001\\ 
    \hline
    SeFT & \RaggedRight lr=0.001, n\_phi\_layers=4, phi\_width=128, phi\_dropout=0.2,
n\_psi\_layers=2, psi\_width=64, psi\_latent\_width=128, dot\_prod\_dim=128,
n\_heads=4, attn\_dropout=0.5, latent\_width=32,
n\_rho\_layers=2, rho\_width=512, rho\_dropout=0.0,
max\_timescale=100.0,
n\_positional\_dims=4 
& \RaggedRight lr=0.00081, n\_phi\_layers=4, phi\_width=128, phi\_dropout=0.2,
n\_psi\_layers=2, psi\_width=64, psi\_latent\_width=128, dot\_prod\_dim=128,
n\_heads=4, attn\_dropout=0.5, latent\_width=32,
n\_rho\_layers=2, rho\_width=512, rho\_dropout=0.0,
max\_timescale=100.0,
n\_positional\_dims=4
\\
    \hline
    InterpNet &ref\_points=96, units=100, input d/o=0.2, rec d/o=0.2, lr=0.001 &ref\_points=192, units=100, input d/o=0.2, rec d/o=0.2, lr=0.001\\
    \hline
    STraTS(ss-) \& I-STraTS(ss-) &d=32, M=2, h=4, d/o=0.2, lr=0.0005 &d=32, M=2, h=4, d/o=0.2, lr=0.001\\
    \hline
    STraTS \& I-STraTS &d=50, M=2, h=4, d/o=0.2, lr=0.0005 &d=50, M=2, h=4, d/o=0.2, lr=0.0005\\
    \bottomrule
    \end{tabular}
    \label{tab:hyper}
\end{table}

\subsection{Implementation Details}
Table \ref{tab:hyper} lists the hyperparameters  used in the experiments for all models for MIMIC-III and PhysioNet-2012 datasets.
All models are trained using a batch size of $32$ with Adam optimizer and training is stopped when sum of ROC-AUC and PR-AUC does not improve for $10$ epochs. 
For pretraining phase using the self-supervision task, the patience is set to $5$ epochs and epoch size is set to $256,000$ samples.
For MIMIC-III dataset, we set the maximum number of time-steps for GRU-D and InterpNet, and the maximum no. of observations for STraTS using the $99^{th}$ percentile for the same. This is done to avoid memory overflow with batch gradient descent.
The deep models are implemented using keras with tensorflow backend. 
For InterpNet, we adapted the official code from \url{https://github.com/mlds-lab/interp-net}. For GRU-D and SeFT, we borrowed implementations from \url{https://github.com/BorgwardtLab/Set_Functions_for_Time_Series}.
The experiments are conducted on a single NVIDIA GRID P40-12Q GPU. Our implementation and data-processing codes for STraTS are available at \url{https://github.com/sindhura97/STraTS}.

\begin{table}[]
\centering
    \caption{Mortality prediction performance on MIMIC-III and PhysioNet-2012 datasets. The results show mean and standard deviation of the metrics after repeating the experiment $10$ times by sampling $50\%$ labeled data each time.}
    \label{tab:pred_perf}
    \begin{tabular}{llccc}
    \toprule
    & &ROC-AUC &PR-AUC &min(Re,Pr) \\
    \midrule
    \multirow{7}{*}{MIMIC-III}
    &GRU &$0.886\pm0.002$ &$0.559\pm0.005$ &$0.533\pm0.007$\\
    &TCN &$0.879\pm0.001$ &$0.540\pm0.004$ &$0.525\pm0.005$\\
    &SAnD &$0.876\pm0.002$ &$0.533\pm0.011$ &$0.515\pm0.008$\\
    &GRU-D &$0.883\pm0.003$ &$0.544\pm0.007$ &$0.527\pm0.005$\\
    &InterpNet &$0.881\pm0.002$ &$0.540\pm0.007$ &$0.516\pm0.005$\\
    &SeFT &$0.881\pm0.003$ &$0.547\pm0.011$ &$0.524\pm0.01$\\
    &STraTS &$\mathbf{0.891\pm0.002}$ &$\mathbf{0.577\pm0.006}$ &$\mathbf{0.541\pm0.008}$\\
    \hline
    \multirow{7}{*}{PhysioNet-2012}
    &GRU &$0.831\pm0.003$ &$0.468\pm0.008$ &$0.465\pm0.009$\\
    
    &TCN &$0.813\pm0.005$ &$0.430\pm0.01$ &$0.433\pm0.009$\\
    &SAnD &$0.800\pm0.013$ &$0.406\pm0.021$ &$0.418\pm0.018$\\
    &GRU-D &$0.833\pm0.005$ &$0.481\pm0.008$ &$0.468\pm0.012$\\
    &InterpNet &$0.822\pm0.007$ &$0.460\pm0.017$ &$0.455\pm0.017$\\
    &SeFT &$0.832\pm0.005$ &$0.454\pm0.017$ &$0.465\pm0.009$\\
    &STraTS &$\mathbf{0.839\pm0.008}$ &$\mathbf{0.498\pm0.012}$ &$\mathbf{0.483\pm0.01}$\\
    \bottomrule
    \end{tabular}
\end{table}

\begin{figure}
    \centering
    \includegraphics[scale=0.415, trim={0.4cm 0 0 0} ]{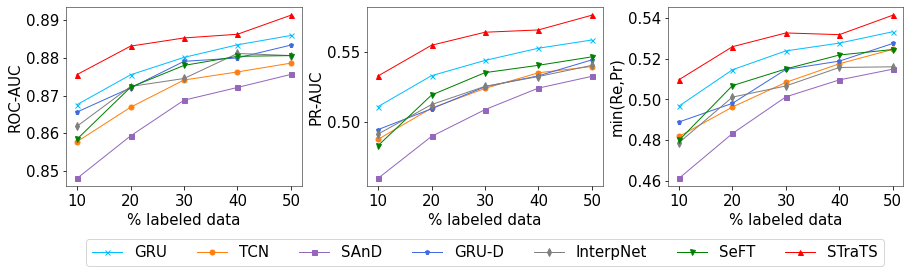}
    \caption{Mortality prediction performance on MIMIC-III dataset for different percentages of labeled data averaged over $10$ runs.}
    \label{fig:gen_mimic_mort}
\end{figure}
\begin{figure}
    \centering
    \includegraphics[scale=0.55, trim={0.5cm 0 0 0} ]{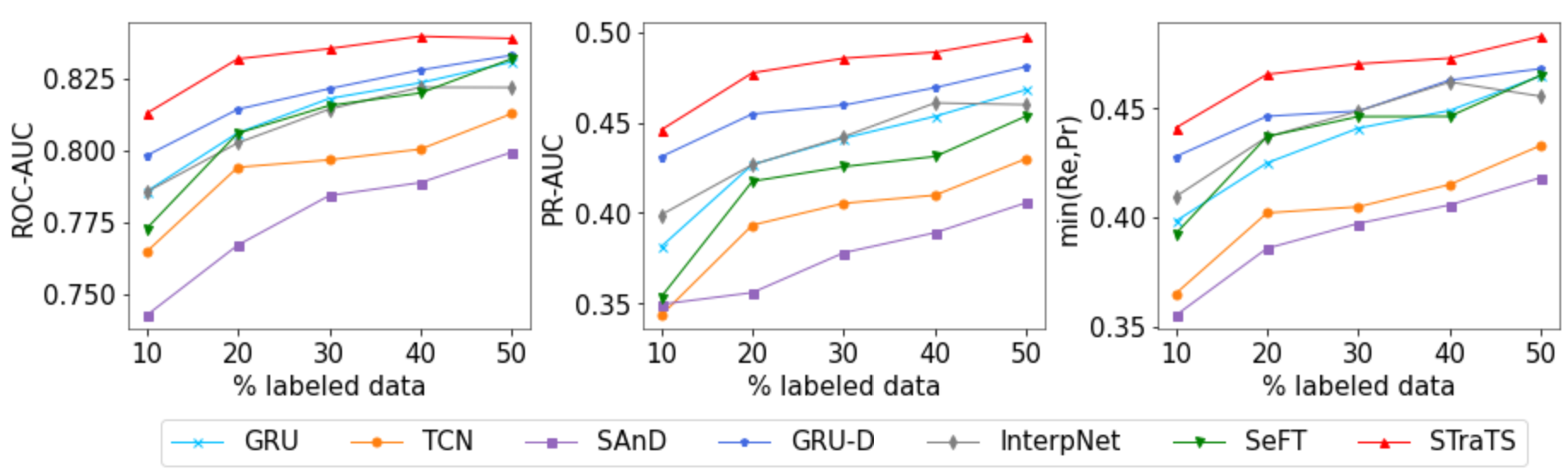}
    \caption{Mortality prediction performance on PhysioNet-2012 dataset for different percentages of labeled data averaged over $10$ runs.}
    \label{fig:gen_phy_mort}
\end{figure}

\subsection{Prediction Performance}
We train each model using $10$ different random samplings of $50\%$ labeled data from the train and validation sets. Note that STraTS uses the entire labeled data and additional unlabeled data (if available) for self-supervision. 
Table \ref{tab:pred_perf} shows the results for mortality prediction on MIMIC-III and PhysioNet-2012 datasets which are averaged over the $10$ runs. 
STraTS achieves the best performance on all metrics, improving PR-AUC by $3.2\%$ and $3.5\%$ on MIMIC-III and PhysioNet-2012 datasets over the best baseline,  respectively. This shows that our design choices of triplet embedding, attention-based architecture, and self-supervision enable STraTS to learn better representations.
We expected the interpolation-based models GRU-D and InterpNet to outperform the simpler models GRU, TCN, and SaND. This was true for all cases except that GRU showed a better performance than GRU-D and InterpNet on the MIMIC-III dataset, for reasons that are unclear.

To test the generalization ability of different models, we evaluate STraTS and the baseline models by training them on varying percentages of labeled data. Lower proportions of labeled data can be observed in real-world when there are several right-censored samples.
Figures \ref{fig:gen_mimic_mort} and \ref{fig:gen_phy_mort} show the results for MIMIC-III and PhysioNet-2012 datasets, respectively. The performance of all models degrades with reduced amount of labeled data. But STraTS is seen to have a crucial advantage compared to other models in scarce labeled data settings which can be attributed to self-supervision.

\subsection{Ablation Study}
We compared the predictive performance of STraTs and I-STraTS, with and without self-supervision and the results are reported in Table \ref{tab:ablation}. `ss+' and `ss-' are used to refer to models trained with and without self-supervision, respectively. We observe that (i) Adding interpretability to STraTS slightly reduces the prediction scores as a result of constraining model representations. (ii) Adding self-supervision improves performance of both STraTS and I-STraTS. (iii) I-STraTS(ss+) outperforms STraTS(ss-) on all metrics on MIMIC-III dataset, and on the PR-AUC metric for PhysioNet-2012 dataset. This demonstrates that the performance drop from introducing interpretability can be compensated by the performance improvements obtained through self-supervision. 
\begin{table}[]
\centering
    \caption{Ablation Study: comparing mortality prediction performance of STraTS and I-STraTS with and without self-supervision. (`ss+' and `ss-' are used to indicate models trained with and without self-supervision, respectively.)}
    \label{tab:ablation}
    \begin{tabular}{llccc}
    \toprule
    & &ROC-AUC &PR-AUC &min(Re,Pr) \\
    \midrule
    \multirow{4}{*}{MIMIC-III}
    &I-STraTS (ss-) &$0.878\pm0.002$&$0.542\pm0.006$&$0.516\pm0.008$\\    
    &I-STraTS (ss+) &$0.887\pm0.003$&$0.556\pm0.008$&$0.531\pm0.005$\\
    &STraTS (ss-)      &$0.881\pm0.002$&$0.546\pm0.007$&$0.525\pm0.012$\\ 
    &STraTS (ss+) &$\mathbf{0.891\pm0.002}$&$\mathbf{0.577\pm0.006}$&$\mathbf{0.541\pm0.008}$\\
    \hline
    \multirow{4}{*}{PhysioNet-2012}
    &I-STraTS (ss-) &$0.826\pm0.008$&$0.456\pm0.018$&$0.467\pm0.025$\\
    &I-STraTS (ss+) &$0.833\pm0.007$&$0.478\pm0.015$&$0.466\pm0.010$\\
    &STraTS (ss-)  &$0.835\pm0.009$&$0.467\pm0.023$&$0.471\pm0.017$\\
    &STraTS (ss+)  &$\mathbf{0.839\pm0.008}$&$\mathbf{0.498\pm0.012}$&$\mathbf{0.483\pm0.010}$\\
    \bottomrule
    \end{tabular}
\end{table}

\subsection{Interpretability}
To illustrate how I-STraTS explains its predictions, we present a case study for an $85$ year old female patient from the MIMIC-III dataset who expired on the $6^{th}$ day after ICU admission. The I-STraTS model predicts the probability of her in-hospital mortality as $0.94$ using only the data collected on the first day. The patient had $380$ measurements corresponding to $58$ time-series variables. The top $5$ variables ordered by their average `contribution score' along with the range (for multiple observations)  or value (for only one observation) are shown in Table \ref{tab:interp}. In addition to old age, we can also observe that I-STraTS considers the abnormal values of Lactate, LDH, Platelet count, and RDW as the most important factors in predicting that the patient is at high risk of mortality. The discharge summary for this patient indicates PEA arrest as the cause of death. Elevated Lactate and LDH levels as seen in this case are known to be associated with cardiac arrest \citep{dell2017prognostic, farhana2021biochemistry}.
Such predictions can not only guide the care givers in identifying high-risk patients for better resource allocation but also guide the clinicians into understanding the contributing factors and make better diagnoses and treatment choices, especially at the early stages of treatment before the condition becomes more severe and uncontrollable.

\begin{figure}
    \centering
    \includegraphics[scale=0.37]{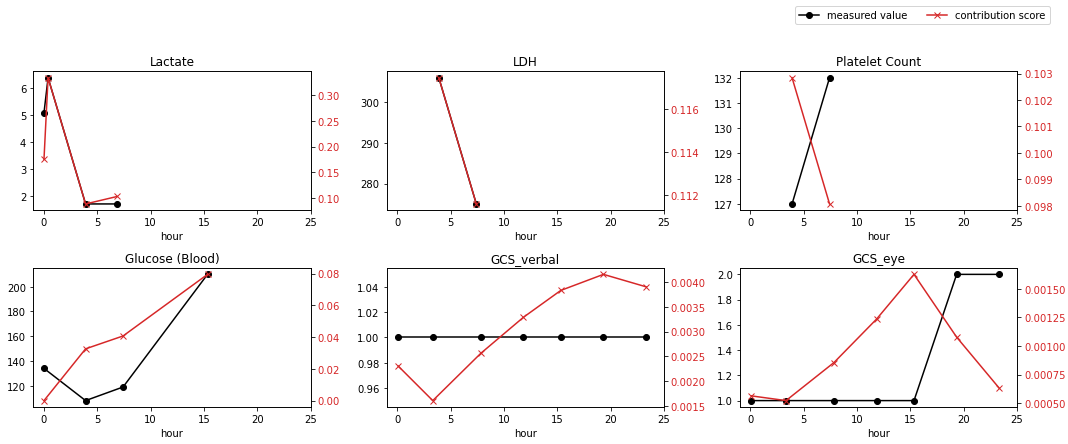}
    \caption{Case study: An illustration of a few time-series with contribution scores for a patient from MIMIC-III dataset.}
    \label{fig:interp}
\end{figure}

To obtain a more fine-grained intuition, the observed time-series for some variables in this ICU stay are plotted in Figure \ref{fig:interp} along with the corresponding contribution scores. It is interesting to see that the contribution scores appear to be positively or negatively correlated with the underlying values or time for several variables. For example, the model gives more weight to higher values of Lactate and LDH that are linked to cardiac arrest which is the patient's cause of death. Similarly, the model pays more attention to increased blood glucose of $210$ mg/dL. As GCS-verbal remains at a constant low of $1$, the model gives it more and more weight as time progresses.

\begin{table}[]
\centering
\caption{Case study: Top 5 variables ordered by `average contribution score' obtained from I-STraTS model for an ICU stay from MIMIC-III dataset.}
\label{tab:interp}
\begin{tabular}{lcc}
\toprule
Variable &Range/Value &`avg. contribution score' \\
\midrule
Age &85 &0.458 \\
Lactate &[1.7, 6.4] mmol/L &0.175 \\
LDH &[275, 306] IU/L &0.115 \\
Platelet Count &[127, 132] K/uL &0.100 \\
RDW &[22.0-22.1]\% &0.083 \\
\bottomrule
    \end{tabular}
\end{table}

\section{Conclusion}
\label{sec:conc}
We proposed a Transformer-based model, STraTS, for prediction tasks on multivariate clinical time-series to address the challenges faced by existing methods in this domain. 
Our approach of using observation triplets as time-series components avoids the problems faced by aggregation and imputation methods for sparse and sporadic multivariate time-series. We used a novel CVE technique which uses parameterized embeddings for continuous values and a multi-head attention to learn contextual representations. 
The self-supervision task of forecasting using unlabeled data enables STraTS to learn more generalized representations, thus outperforming state-of-the-art baselines. In addition, we also showed that STraTS generalizes well even when labeled data is scarce and is also more robust to noise compared to existing methods. 
We also proposed an interpretable version of STraTS, called I-STraTS, for which self-supervision compensates the drop in prediction performance from introducing interpretability. This work can motivate other researchers to explore more self-supervision tasks for clinical time-series data. Along with exploring more self-supervision tasks, future work should look at adapting STraTS or optimizing its computational efficiency for longer time series where attention matrices can become large and infeasible.


\begin{acks}
We thank Lakshmi Tipirneni for her help with the clinical domain knowledge related to the extraction of our time-series dataset from MIMIC-III database and for providing clinical insights on the case study presented in Section 4.7. This work was supported in part by the US National Science Foundation grants IIS-1838730 and Amazon AWS credits.

\end{acks}

\bibliographystyle{ACM-Reference-Format}
\bibliography{6.references}


\end{document}